*Ососков Г.А., Дмитриевский С.Г., Стадник А.В*

## НЕЙРОННЫЕ СЕТИ С САМООРГАНИЗАЦИЕЙ В ЗАДАЧАХ КЛАССИФИКАЦИИ И ОБРАБОТКИ ИЗОБРАЖЕНИЙ

*Объединенный институт ядерных исследований,
Дубна, Московской обл.,email: ososkov@jinr.ru*

**1. Введение.** Достижения теоретической физики и промышленной электроники последних десятилетий, вызвали настоящий переворот в экспериментальной физике высоких энергий (ФВЭ), в которой произошел важный переход к коллайдерным ускорителям на встречных пучках в Тэв-ных областях энергий и к современным электронным детекторам, способным регистрировать гигантские потоки экспериментальных данных. Их обработка неизбежно потребовала привлечения новых математических средств таких, в частности, как искусственные нейронные сети (ИНС), клеточные автоматы (КА) и вейвлет-анализ для распознавания треков заряженных частиц, проверки физических гипотез и быстрой обработки изображений.

Настоящая работа посвящена некоторым из новых подходов в этой области, развитым в Объединенном институте ядерных исследований, на примерах их применений к двум важным прикладным задачам: локализация эмульсионного «кирпича» с вершиной взаимодействия в эксперименте OPERA по поиску осцилляций нейтрино и считывание номеров проезжающих автомобилей в режиме реального времени.

**2. Об используемых типах нейронных сетей.** В этой работе мы применяем **прямоточные ИНС без обратных связей,** подразделяемые на два класса: с управляемым обучением, например, многослойные персептроны (МСП) или радиально-базисные сети (РБС) и сети естественной классификации, такие как **самоорганизующаяся карта Кохонена**, в которой выходные нейроны расположены обычно в виде двухмерной решетки, позволяющей конфигурировать входной образ в соответствии с топологическим представлением исходных данных. Мы используем также и **клеточные автоматы** как специальный тип нейронных сетей с локальными связями. На предыдущей IV-й конференции в Дивноморске, а также в работах [1-2] мы предложили новый тип нейронной сети, предназначенной для решения задачи классификации со свойствами самоорганизации в процессе обучения. Такая **самонастраивающаяся радиально-базисная нейронная сеть** (СНРБ-сеть) показала показала многообещающие возможности как в известных сложных задачах классификации типа распутывания двух скрученных спиральных множеств точек, так и в практической задаче распознавания человеческих лиц.

Алгоритм обучения СНРБ-сети (см [2]), позволяет существенно сократить количество нейронов в скрытом промежуточном слое, увеличивая скорость ее работы, поскольку ее обучение одновременно является и процессом построения её самой. В результате процесса обучения СНРБ-сети определяются не только значения синаптических весов нейронов, но и количество нейронов в слоях и даже количество слоев. Результирующая комбинация состоит из трех или четырех слоев нейронов (или то же самое, двух или трех слоев синаптических весов) в зависимости от обучения, и может быть дополнена слоем реализующем метод главных компонент, в зависимости от типа данных в задаче.

**3. Классификация событий и поиск вершины в эксперименте OPERA.** Одной из важных проблем экспериментальной науки является обнаружение массы нейтрино, что позволило бы сделать прорыв во многих областях современной физики. В настоящий момент для поиска процесса осцилляций нейтрино $v_\mu \leftrightarrow v_\tau$, который подтвердил бы наличие массы у этой элементарной частицы, готовится международный эксперимент OPERA. Поиск взаимодействия $v_\tau$ с веществом

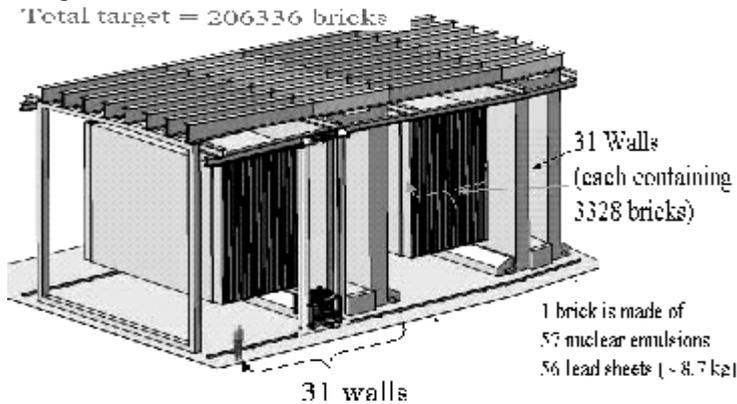

Рис.1 Общий вид установки OPERA

выполняется в мишенном трекере, образованном 31 стенкой, каждая из которых составлена из 3328 «кирпичей», состоящих из слоев фотоэмульсии и свинца. Стенки разделяются слоями сцинтиляционных полос, позволяющих грубо определять координаты кирпичей. После того, как в детекторе зарегистрировано какое-то событие, необходимо идентифицировать ту стенку и тот кирпич эмульсии, в котором расположена вершина реакции взаимодействия нейтрино с веществом. Эта идентификация существенно затрудняется из-за наличия шумовых отсчетов и, главное, отсчетов, вызванных процессом обратного рассеивания. В работе [3] для целей идентификафии были успешно применен нейросетевой подход. Поиск нужной стенки, а затем и кирпича с вершиной взаимодействия осуществлялся с помощью трехслойных персептронов, обучаемых отдельно на разных классах событий, смоделированных для разных физических каналов взаимодействия. Нами предложены и реализованы два пути повышения эффективности идентификации: (1) проведение предварительной более естественной классификации событий, возникающих в детекторе OPERA, с помощью сети Кохонена; (2) замена МСП на СНРБ-сеть. Первые результаты расчетов показывают плодотворность нашего подхода.

**4. Распознавание номера автомобиля по изображению.** Алгоритм распознавания номера автомобиля состоит из двух достаточно независимых этапов: (1) нахождение и выделение части изображения, содержащей номер; (2) непосредственное распознавание номера.

**Первый этап** выполняется с помощью **вейвлет-преобразование** изображения [4]. Постановка задачи с распознаванием номеров автомобилей, которые являются двумерными объектами, выглядит намного сложнее, чем вейвлет-фильтрация одномерных спектров. Поэтому мы, прежде всего, сократили размерность задачи, зафиксировав масштаб вейвлет-преобразования в значении, определяемом размерами области номера, частотой чередования черного и белого в этой области и порядком выбранного вейвлета. Например, для гауссова вейвлета 6-го порядка $g_6$ (см.[5]) подходит масштаб равный 2, т.к. это дает около 12-ти пересечений нуля, что достаточно для описания номера.

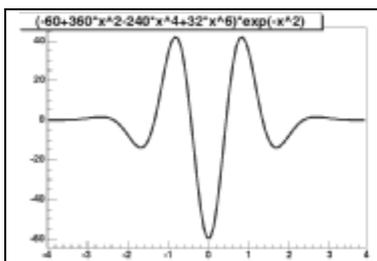

Рис.2. Гауссов вейвлет шестого порядка $g_6$

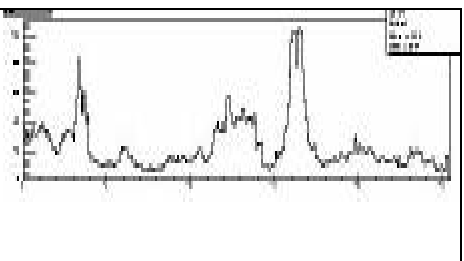

Рис.3. Пример линии вейвлета $g_6$ для изображений с рис. 4а

Выполняя вейвлет-преобразование с фиксированным порядком каждой строки исходного изображения, мы получаем, таким образом, не двумерный, а одномерный спектр. При этом суммарное представление результатов $g_6$-преобразования для всех строк дает нам двумерный вейвлет-домен, в котором область больших вейвлет-коэффициентов, соответствующая чередованию черных и белых полос в области номера, представлена в виде характерного плато, симметричного и с крутыми склонами с обеих сторон. Эта крутизна границ области вейвлет-коэффициентов, соответствующей номеру может быть детектирована по большим значениям второй производной. Однако вместо сложной с вычислительной точки зрения процедуры вычисление вторых производных для каких-то двумерных подобластей растровых изображений мы предпочли использовать метод **бинаризации вейвлет-пикселей** с последующим применением **специального клеточного автомата** (КА) для кластеризации. При бинаризации все пиксели вейвлет-изображения упорядочиваются по уровням серого. Затем нижние 70% становятся черными, а верхние 30% - белыми. Дисциплина выживания каждой белой клетки КА в зависимости от восьми ее соседей, организована так, чтобы «выживала» слитная группа единичных пикселей, соответствующая области номера, а окружающие её отдельные единицы вымирали. Эффективность предложенного метода выделения области изображения с неким номером продемонстрирована на его применении к гораздо более сложной задаче по распознаванию номеров на товарных вагонах.

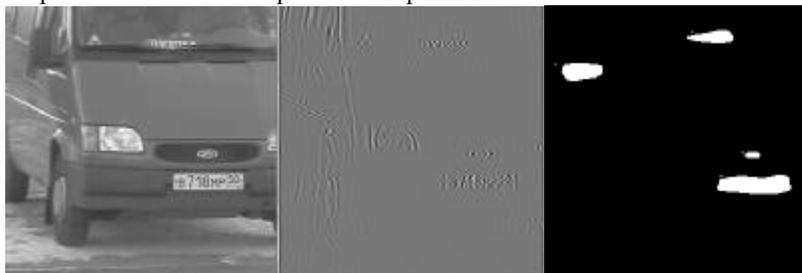

Рис.4. (слева направо) (а) изображение автомобиля, (б) вид $g_6$ –преобразования, выполненного для каждой строки развертки этого изображения, (в) результат работы клеточного автомата.

**Второй этап.** После контрастирования фрагмента изображения с номером проводилось разбиение номера на отдельные цифры и его распознавание СНРБ-сетью, обученной на предварительно «размытых» образцах известного шрифта Lucida typewriter программы MsWord.

1. G.Ososkov, A.Stadnik, Effective training algorithms for RBF-neural networks, *Nucl. Instr. Meth.* **A502/2-3 (2003) 529-531.**
2. Г.А.Ососков, А.В.Стадник, Эффективные нейросетевые алгоритмы для обработки экспериментальных данных, *ИТВС, №1 103-125, УРСС, Москва, 2004.*